\documentclass[
twocolumn,
]{ceurart}

\usepackage{listings}
\usepackage[draft=false]{defs-common}
\usepackage{defs-custom}

\begin{document}

\copyrightyear{2022}
\copyrightclause{Copyright 2022 for this paper by its authors. Use permitted under Creative Commons License Attribution 4.0 International (CC BY 4.0).}

\conference{The Joint Workshop Proceedings of the 2022 Conference on Artificial Intelligence and Interactive Digital Entertainment}

\title{Level Assembly as a Markov Decision Process}

\author[1]{Colan F. Biemer}[email=biemer.c@northeastern.edu]
\author[1]{Seth Cooper}[email=se.cooper@northeastern.edu]
\address[1]{Northeastern University, Boston, MA, USA}

\begin{abstract}
Many games feature a progression of levels that doesn't adapt to the player. This can be problematic because some players may get stuck if the progression is too difficult, while others may find it boring if the progression is too slow to get to more challenging levels. This can be addressed by building levels based on the player's performance and preferences. In this work, we formulate the problem of generating levels for a player as a Markov Decision Process (MDP) and use adaptive dynamic programming (ADP) to solve the MDP before assembling a level. We tested with two case studies and found that using an ADP outperforms two baselines. Furthermore, we experimented with player proxies and switched them in the middle of play, and we show that a simple modification prior to running ADP results in quick adaptation. By using ADP, which searches the entire MDP, we produce a dynamic progression of levels that adapts to the player.
\end{abstract}


\maketitle

\newcommand{\shortcite}[1]{\cite{#1}}


\newcommand{\XFIGUREplaceholder}{
\begin{figure}[t]
\centering
\includegraphics[width=0.975\columnwidth]{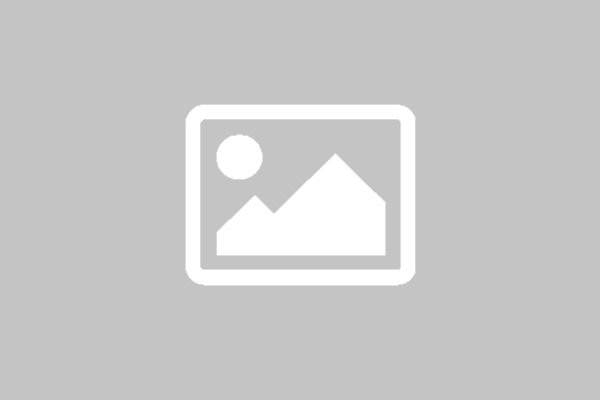}
\caption{\label{XFIGUREplaceholder} Placeholder figure.}
\end{figure}
}

\newcommand{\XFIGUREngramrewardplots}{
\begin{figure*}
    \centering
    \includegraphics[width=.9\textwidth]{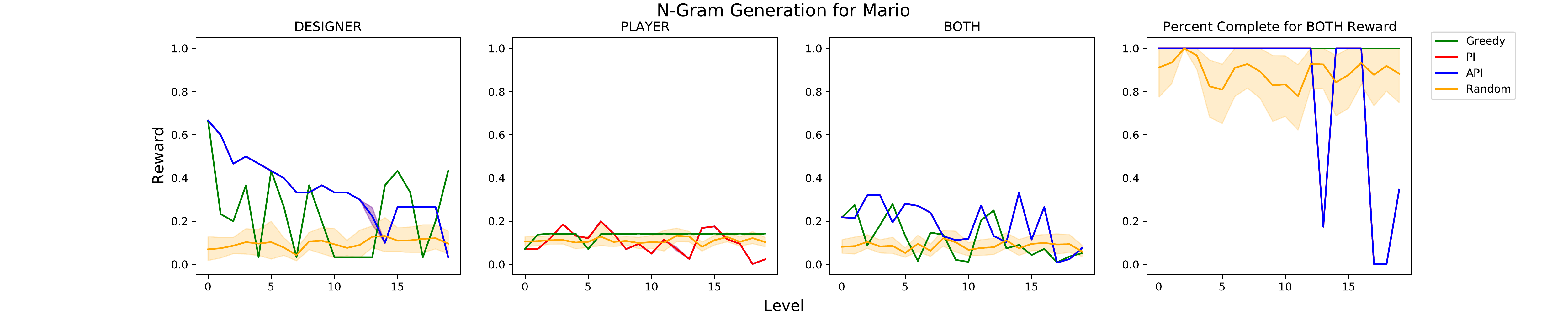}
    \caption{Average reward for all three reward types when running n-gram generation for \textit{Mario}, and shows the average percent completable for the \textit{both} reward.}
    \label{fig:XFIGUREngramrewardplots}
\end{figure*}
}

\newcommand{\XFIGUREsegmentsgraph}{
\begin{figure*}
    \centering
    \includegraphics[width=.8\textwidth]{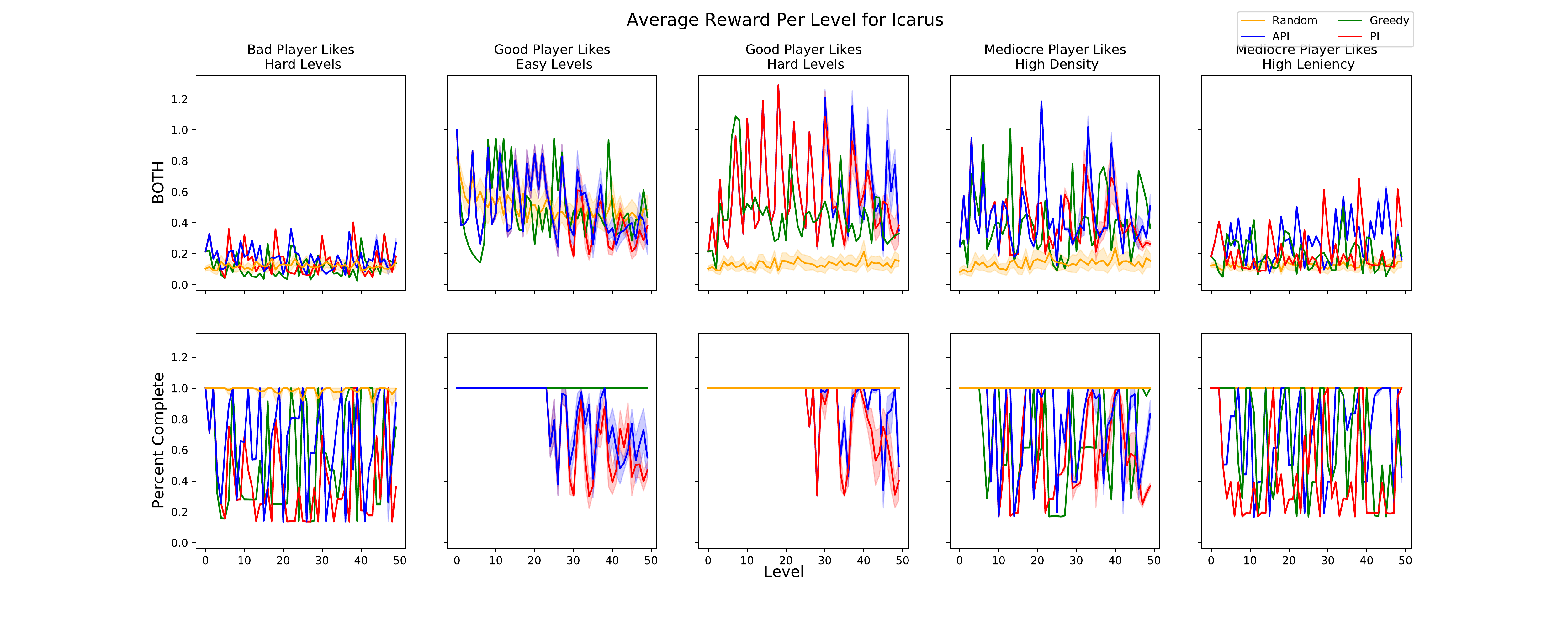}
    \caption{The first three rows show the average reward for all three reward types for five player proxies using level segments built for \textit{Icarus}. The final rows shows percent completable per level when using the \textit{both} reward. }
    \label{fig:XFIGUREsegmentsgraph}
\end{figure*}
}

\newcommand{\XFIGUREicarusheatmaps}{
\begin{figure*}
    \centering
    \begin{tabular}{c}
    \includegraphics[width=0.9\textwidth]{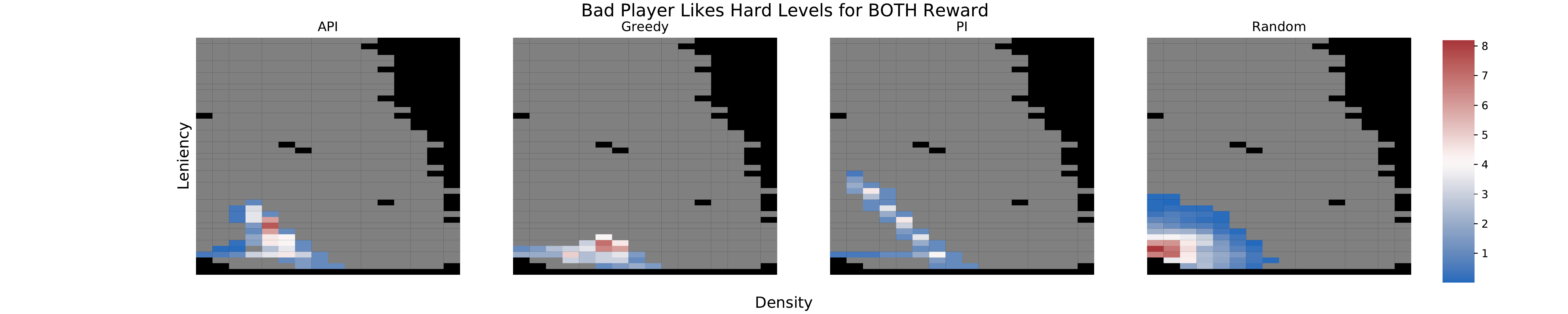}
    \\
    \includegraphics[width=0.9\textwidth]{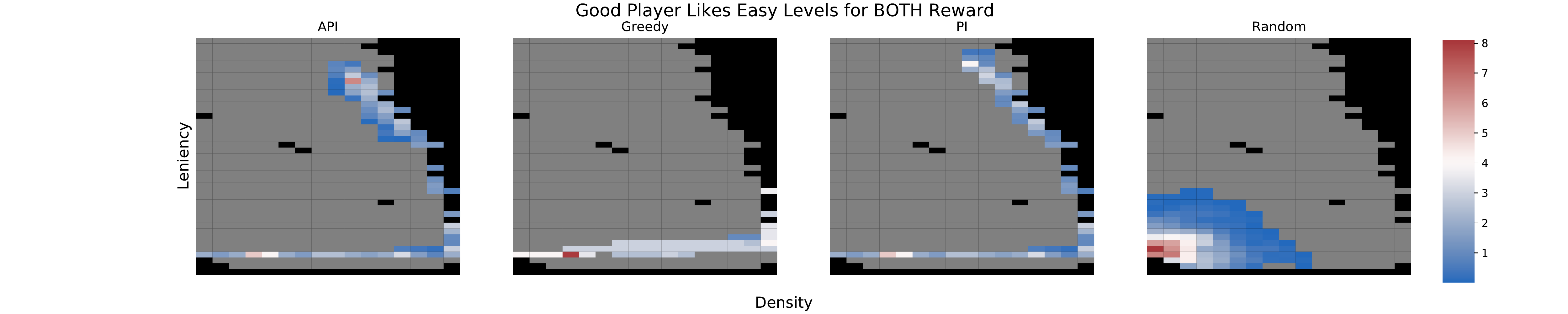}
    \\
    \includegraphics[width=0.9\textwidth]{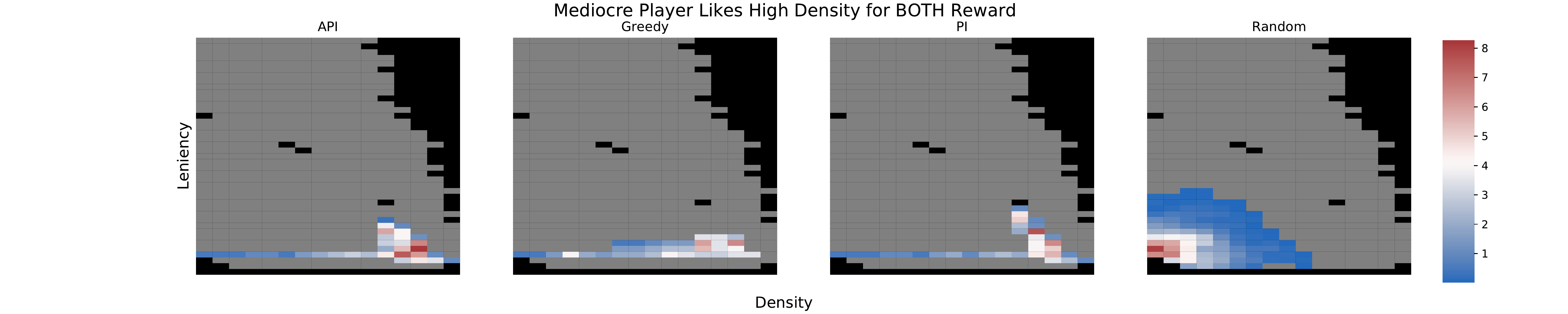}
    \end{tabular}
    
    \caption{Heat map of cells in the MAP-Elites grid visited by each director for \textit{Icarus} segment generation.}
    \label{fig:XFIGUREicarusheatmaps}
\end{figure*}
}

\newcommand{\XFIGUREheatmapmediocreplayerlikeshighdensity}{
\begin{figure*}
    \centering
    \includegraphics[width=.9\textwidth]{figure/segment_heatmap_Mediocre Player Likes High Density_BOTH.pdf}
    \caption{Heat map of cells in the MAP-Elites grid visited by each director for \textit{Icarus} segment generation when using both rewards and the player proxy Mediocre Player Likes High Density.}
    \label{fig:XFIGUREheatmapmediocreplayerlikeshighdensity}
\end{figure*}
}

\newcommand{\XFIGUREheatmapmediocreplayerlikeshighleniency}{
\begin{figure*}
    \centering
    \includegraphics[width=.9\textwidth]{figure/segment_heatmap_Mediocre Player Likes High Leniency_BOTH.pdf}
    \caption{Heat map of cells in the MAP-Elites grid visited by each director for \textit{Icarus} segment generation when using both rewards with \MPLHL}
    \label{fig:XFIGUREheatmapmediocreplayerlikeshighleniency}
\end{figure*}
}

\newcommand{\XFIGUREheatmapbadplayerlikeshardlevels}{
\begin{figure*}
    \centering
    \includegraphics[width=.9\textwidth]{figure/segment_heatmap_Bad Player Likes Hard Levels_BOTH.pdf}
    \caption{Heat map of cells in the MAP-Elites grid visited by each director for \textit{Icarus} segment generation when using both rewards for \BPLHL.}
    \label{fig:XFIGUREheatmapbadplayerlikeshardlevels}
\end{figure*}
}

\newcommand{\XFIGUREheatmapgoodplayerlikeseasylevels}{
\begin{figure*}
    \centering
    \includegraphics[width=.9\textwidth]{figure/segment_heatmap_Good Player Likes Easy Levels_BOTH.pdf}
    \caption{Heat map of cells in the MAP-Elites grid visited by each director for \textit{Icarus} segment generation when using both rewards and the player proxy \textit{Good Player Likes Easy Levels}.}
    \label{fig:XFIGUREheatmapgoodplayerlikeseasylevels}
\end{figure*}
}

\newcommand{\XFIGUREswitchingplayerplots}{
\begin{figure}
    \centering
    \includegraphics[width=.37\textwidth]{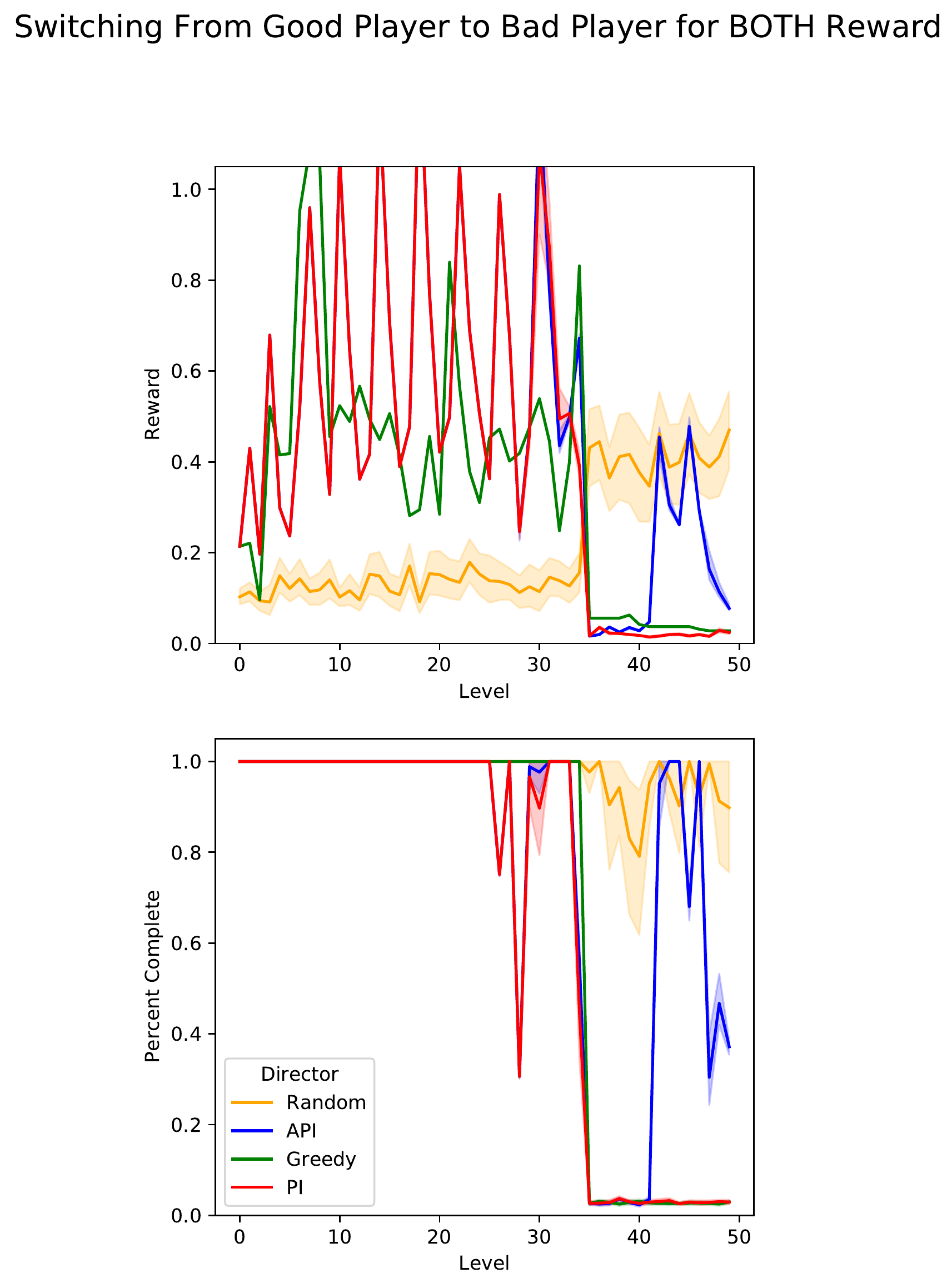}
    \caption{Shows the average reward per level (top) and the average percent complete average per level (bottom) when the player is switched after 35 levels from \GPLHL{} to \BPLEL{}.}
    \label{fig:XFIGUREswitchingplayerplots}
\end{figure}
}

\newcommand{\XFIGUREicaruslevel}{
\begin{figure*}
    \centering

    \begin{tabular}{lc}
        \rotatebox[origin=r]{90}{Level 5\:\:\:\:\:} 
        & \includegraphics[angle=270,width=.8\textwidth]{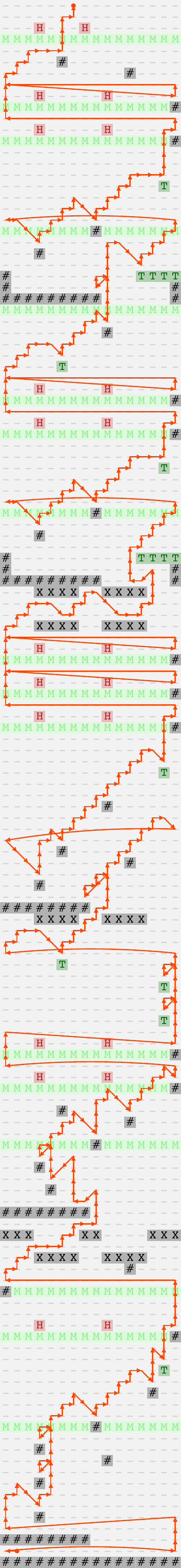} 
        \\
        \rotatebox[origin=r]{90}{Level 6\:\:\:\:\:}
        &
        \includegraphics[angle=270,width=.8\textwidth]{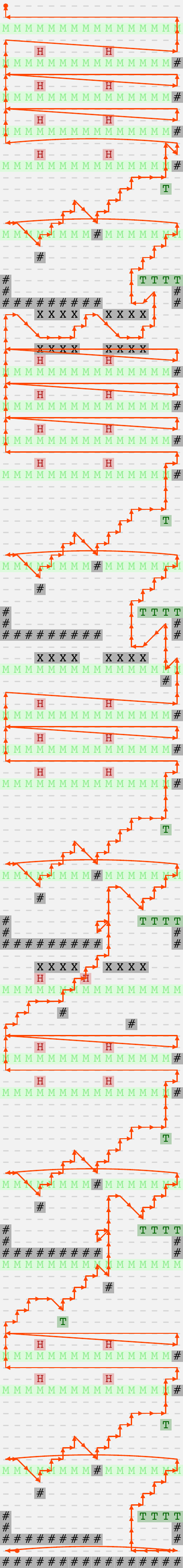}
    \end{tabular}
    \caption{Two \textit{Icarus} levels which have been generated by PI for \GPLHL. Levels are rotated for visibility and the red arrows show the path taken by an agent.}
    \label{fig:XFIGUREicaruslevel}
\end{figure*}
}

\newcommand{\XFIGUREswitchingplayerheatmap}{
\begin{figure*}
    \centering
    \includegraphics[width=.9\textwidth]{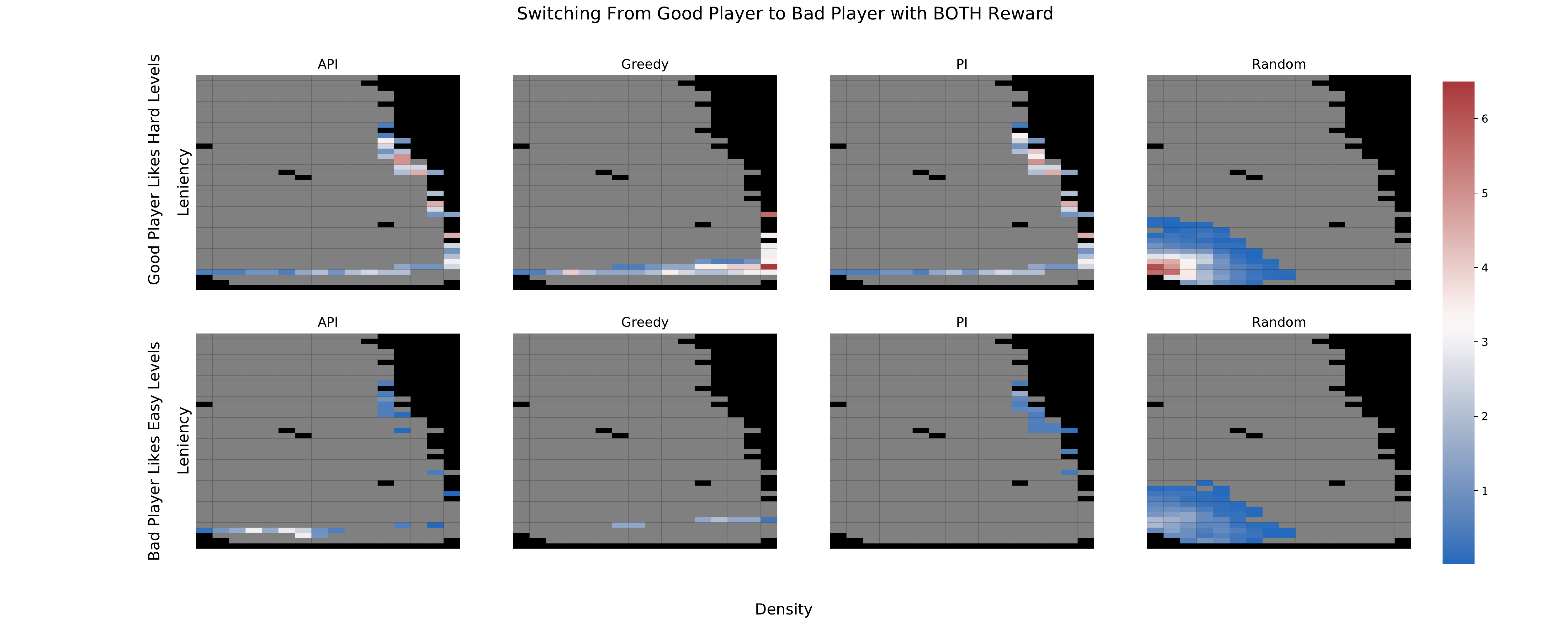}
    \caption{Heat map of cells in the MAP-Elites grid visited by each director for \textit{Icarus} segment generation when switching players after 35 levels generated. The second proxy is run for 15 levels.}
    \label{fig:XFIGUREswitchingplayerheatmap}
\end{figure*}
}

\newcommand{\XTABLEPlayerPersona}{
\begin{table*}
\centering
\begin{tabular}{|l|c|c|c|}
\hline
 Player Proxy & Always Wins Threshold & Fail Percent Complete & Player Reward 
\\
\hline
\BPLHL & $D+L < 0.25\:MAX\_BC$ & 0.25 - 0.40 & $BC/MAX\_BC$ \\
\BPLEL & $D+L < 0.25\:MAX\_BC$ & 0.25 - 0.40  & $1 - BC/MAX\_BC$ \\
\MPLHD & $D+L < 0.50\:MAX\_BC$ & 0.50 - 0.70 & $D$ \\
\MPLHL & $D+L < 0.50\:MAX\_BC$ & 0.50 - 0.70 & $L$ \\
\GPLHL & $D+L < 0.75\:MAX\_BC$ & 0.75 - 0.95 & $BC/MAX\_BC$ \\
\GPLEL & $D+L < 0.75\:MAX\_BC$ & 0.75 - 0.95 & $1 - BC/MAX\_BC$ \\ 
\hline
\end{tabular}
\caption{\label{XTABLEPlayerPersona} Player proxies used to assess level segments. $D$ stands for density and $L$ for leniency. $MAX\_BC$ represents the maximum sum of the BCs, in the case of \textit{Icarus} it is 2. Fail Percent Complete represents the range of percent complete values used if the level is greater than the always win threshold. Player Reward is a substitute for a player model, $M(s)$.}
\end{table*}
}

\newcommand{\XTABLENGramRewards}{
\begin{table}
\centering
\begin{tabular}{|l|c|c|c|}
\hline
Director & Reward & Percent Complete
\\
\hline
API & $\mathbf{0.1878 \pm 0.0976}$ & $0.7912 \pm 0.3623$ \\
PI & $\mathbf{0.1878 \pm 0.0976}$ & $0.6763 \pm 0.4453$ \\
Greedy & $0.1172 \pm 0.0884$ & $\mathbf{1.0000 \pm 0.0000}$ \\
Random & $0.0864 \pm 0.0843$ & $0.9798 \pm  0.1309$ \\
\hline
\end{tabular}
\caption{\label{XTABLENGramRewards} Average plus or minus the standard deviation of the reward and percentage completed for each director for Case Study 1 on n-grams for \textit{Mario} level assembly.}
\end{table}
}

\newcommand{\XTABLEIcarusRewards}{
\begin{table*}
\centering
\begin{tabular}{|l |l|c|c|c|}
\hline
Player & Director & Reward & Percent Complete
\\
\hline
\multirow{3}{6em}{\BPLEL} 
 & API & $\mathbf{0.1698 \pm 0.0767}$ & $0.6555 \pm 0.2986$ \\
 & PI & $0.1522 \pm 0.0912$ & $0.5594 \pm 0.3368$ \\
 & Greedy & $0.1178 \pm 0.0739$ & $0.4046 \pm 0.2770$ \\
 & Random & $0.1208 \pm 0.0814$ & $\mathbf{0.9884 \pm  0.0808}$ \\
\hline
\multirow{3}{6em}{\GPLEL} 
 & API & $\mathbf{0.5079 \pm 0.2486}$ & $0.8442 \pm 0.2683$ \\
 & PI & $0.4826 \pm 0.2519$ & $0.7870 \pm 0.3043$ \\
 & Greedy & $0.4965 \pm 0.2216$ & $\mathbf{1.0000 \pm 0.0000}$ \\
 & Random & $0.5009 \pm 0.2143$ & $\mathbf{1.0000 \pm 0.0000}$ \\
\hline
\multirow{3}{6em}{\GPLHL} 
 & API & $\mathbf{0.6093 \pm 0.3150}$ & $0.9217 \pm 0.2053$ \\
 & PI & $0.5751 \pm 0.3070$ & $0.8639 \pm 0.2716$ \\
 & Greedy & $0.4519 \pm 0.2006$ & $\mathbf{1.0000 \pm 0.0000}$ \\
 & Random & $0.1323 \pm 0.0972$ & $\mathbf{1.0000 \pm 0.0000}$ \\
\hline
\multirow{3}{6em}{\MPLHD} 
 & API & $\mathbf{0.4427 \pm 0.2320}$ & $0.76840 \pm 0.2964$ \\
 & PI & $0.3974 \pm 0.2183$ & $0.6278 \pm 0.3028$ \\
 & Greedy & $0.4187 \pm 0.2140$ & $0.6847 \pm 0.3389$ \\ 
 & Random & $0.1375 \pm 0.1205$ & $\mathbf{1.0000 \pm 0.0000}$  \\
\hline
\multirow{3}{6em}{\MPLHL} 
 & API & $\mathbf{0.2835 \pm 0.1366}$ & $0.7089 \pm 0.2886$ \\
 & PI & $0.2112 \pm 0.1422$ & $0.4494 \pm 0.3131$ \\
 & Greedy & $0.1950 \pm 0.1015$ & $0.6609 \pm 0.3229$ \\
 & Random & $0.1272 \pm 0.0858$ & $\mathbf{1.0000 \pm 0.0000}$ \\
\hline
\multirow{3}{6em}{All Players} 
 & API & $\mathbf{0.4026 \pm 0.2694}$ & $0.7797 \pm 0.2895$ \\
 & PI & $0.3681 \pm 0.2695$ & $0.6269 \pm 0.3518$ \\
 & Greedy & $0.3317 \pm 0.2285$ & $0.7810 \pm 0.3092$ \\
 & Random & $0.2038 \pm 0.1972$ & $\mathbf{0.9977 \pm 0.0364}$ \\
\hline
\end{tabular}
\caption{\label{XTABLEIcarusRewards} Average plus or minus the standard deviation of the reward and percentage completed for each director for Case Study 2 for \textit{Icarus} level assembly.}
\end{table*}
}


\section{Introduction}

Many games on the market today feature a form of static level progression (that is, levels are played in a fixed order). An alternative is a level progression that adapts to the player. At face value, such a dynamic approach seems preferable. So, why not use it? First, we need to take into account the workload. A static progression has a minimal number of levels required. A dynamic progression requires multiple levels for different players, requiring more work from already overworked developers \cite{danastasio_2021}. A potential solution is procedural content generation \cite{shaker2016procedural}, but designers will still need to do additional work \cite{karth2019addressing}. The second problem is adapting to the player. We could use a player model \cite{yannakakis2013player, hooshyar2018data} to inform level generation. But the resulting system can be complex and may not have the accuracy guarantees that a designer needs to put the system into production. Overall, creating a dynamic progression has a barrier to entry that is too high for most games.  

One of the first cases for adaptive games comes from Hunicke \shortcite{hunicke2005case}, which argues for dynamic difficulty adjustment (DDA). They show how small adjustments and interventions can improve player performance. Jennings-Teats et al. \shortcite{jennings2010polymorph} showed a different approach where a generator adapts levels for a player using models trained before play. The problem with models that do not update with the player in real-time is that they may not generalize to all players. Alternatively, online learning is where a model trains in real time. In our opinion, online learning is a more promising approach for dynamic progressions because it addresses the weaknesses of pre-built models, but there are problems with how quickly these algorithms can adapt to the player. 

In our work,\footnote{\url{https://github.com/crowdgames/mdp-level-assembly}} we formulate the problem of dynamic progressions as a Markov Decision Process (MDP). (See the Approach section for details on MDPs.) We show how to make an MDP from a directed graph of portions of levels that we append together, where each portion of a level is a state in the MDP. The \textit{director} is the agent that makes decisions about how to assemble levels with the MDP. The rewards of the MDP are initialized via a custom reward function that represents designer preference. The reward table updates based on a dynamic reward function that takes into account designer and player preference as well as how many times the player has played a given state. Before assembling the next level, the director can learn from the updated MDP. In this work, we use policy iteration, which is a method in the family of adaptive dynamic programming (ADP). The benefit of using ADP, which is typically an offline approach, is that we use it in an online environment to effectively learn the best possible level formation before each playthrough. 

We test with four directors: Random, Greedy, Policy Iteration (PI), and Adaptive Policy Iteration (API)---a simple modification to PI that we describe below. We apply these directors to two case studies. For the first case study, we initialize an MDP with an n-gram built with \textit{Mario} levels with rewards based on the presence of an enemy in the level slices. An A* agent tests completability and a surrogate player model rewards level slices with at least one enemy. We find that API and PI perform equally well in terms of reward, but API produces levels that are on average more completable. Random is the best performer in terms of completable levels but worst for reward. Our second case study uses an MDP that connects \textit{Kid Icarus} level segments to form a level. We use player proxies \cite{liapis2015procedural, green2018generating} to evaluate level segments, where a proxy has a distribution over the levels it can beat and where it struggles. A proxy also specifies the kind of level segments it enjoys as a player reward. We find that API performs best on average across all the player proxies. Additionally, we test how well directors adapt when a player proxy changes. We find that only API adapts to the change. This lends credence to our argument that ADP with a simple modification, in addition to being straightforward to implement, is an effective method for adapting to players because it explores the whole space, improves data efficiency, and is a promising area for future work in dynamic progressions.

\section{Related Work}


DDA can be achieved through level generation, and Jennings-Teats et al. \shortcite{jennings2010polymorph} gave an early example with \emph{Polymorph}. They used a generator \cite{smith2009rhythm} that generated levels based on rhythms (run, jump, or wait). They collected data with this generator, where players played short levels; their playthroughs and results were stored, as well as the player's subjective difficulty ranking. This data was used to train two models to classify difficulty and player skill. By adopting a generate-and-test approach, the game adapted to the player.



One approach to address the limitations of online-learning techniques comes from Gonzalez-Duque et al. \shortcite{gonzalez2020finding}. In their work, they used MAP-Elites \cite{mouret2015illuminating} to generate a grid of levels for multiple agents, where the objective is to find a level with a sixty percent win rate for the target agent. With these grids, they tested how quickly a grid can be updated for a new agent. They achieved this with the intelligent trial-and-error algorithm \cite{cully2015robots}. In the ideal case, they can adapt to a new agent in one iteration. Overall, this is very similar to our work. We both use the MAP-Elites grid as a starting point and an offline reinforcement learning method in an online environment. However, unlike our work, when working with real players, they need an agent that can reasonably approximate the player for an update, whereas our work does not rely on a surrogate for the player.

One of the earliest formulations of a game as an MDP comes from Thue and Bulitko \shortcite{thue2012procedural} with the procedural game adaption (PGA) framework. The MDP decides what happens next in the world for a player. Outside of the MDP, there is the \textit{manager}, which estimates the player's reward function and policy. With these two models, the manager updates the MDP. Thus, player decisions affect the MDP, which is optimized in real-time to give the player the best experience. Thue and Bulitko did not directly apply PGA to DDA and instead focused on narrative direction and player preferences (e.g. weapon preference).


Shu et al. \shortcite{shu2021experience} built a framework called experience-driven procedural content generation via reinforcement learning, which is composed of four systems: (1) game-playing agent, (2) latent space generator (e.g. a GAN), (3) level repairer, and (4) experience model. This framework is designed after an MDP. A state is a level. An action is a latent vector that is input to a GAN to get a level. The transition model is modeled by a neural network, and it outputs a new state given the previous state as input. The reward is defined by a player experience model.

\section{Approach}

\subsection{Level Assembly as a Markov Decision Process} 
\label{sec:level_generation}

For a diagram of the system, see Appendix A. Our system starts by \textit{initializing the MDP}. A Markov Decision Process (MDP) is a framework for modeling decision-making for discrete time and state environments. In this work, we use an MDP to build a level assembly policy that is tailored to the player. An MDP is made up of a set of states ($s \in S$), actions ($A(s) \in A$), rewards ($R(s) \in R$), and a transition model ($P(s'|s,a) \in P$). In our case, we represent the problem as a graph where a state is equivalent to a node and an action to an edge. Therefore, the transition model is better represented as $P(s'|s,s_a)$ where $s$ is the starting state, $s_a$ is the target state for the action taken, and $s'$ is the state actually traversed to. A state $s$ represents a portion of a level, be it a single slice or a level segment. We include two additional states: $start$ and $death$. The death state is a terminal node with a reward of $-1$. All other states have one to many outgoing edges. Each action can result in the player beating and entering the target state or the player losing and entering the death state. We initialize the probability of the player successfully beating the target state to $P(s_a|s,s_a) \leftarrow 0.99$ and $P(death|s,s_a) \leftarrow 0.01$. Note that choosing the death state is not a valid action. In the case studies below, we discuss each MDP in detail, but for now assume the existence of an MDP. In addition to $R(s)$, we have a static rewards table $R_D(s)$, which is the designer rewards table. This table is built by the designer assigning a reward for every level segment $s$; we use metrics that can be automated, such as counting the number of enemies in a level slice. To initialize $R(s)$ we use $\forall s \in S: R(s) \leftarrow R_D(s)$. In this scheme, the MDP starts with a bias completely toward the designer and, as we show below, updates over time.

The second step is to \textit{build the policy} for level assembly. Given an MDP, a policy $\pi(s)$ chooses a connecting state given the current state as input. Our goal is to find an optimal policy that balances the designer's and the player's reward---we discuss how a policy is built in the Directors section.

The third step in our system is to \textit{generate a level} with the policy. Imagine we want to build a level three states long. We start with $\pi(start)$ and get $s_1$. We input $s_1$ into the policy and get $s_2$. Lastly, we enter $s_2$ and get $s_3$. The assembled level is the concatenation of $s_1$, $s_2$, and $s_3$.

Next, the \textit{player plays the level}.  For this example, let's assume that the player completed $s_1$, failed fifty percent of the way through $s_2$, and never reached $s_3$. At this point, the player can \textit{exit} or choose to play again. In the latter case, we \textit{update the MDP} based on their results.

First, we modify the MDP by updating the neighbors of the $start$ node with any level segment the player beat. In this case, the player only beat $s_1$. If $s_1$ isn't a neighbor of the start node, the edge $(start, s_1)$ is added---probabilities are handled below. In this case, we know $s_1$ is already a neighbor of the start node because the policy selected $s_1$ from the start node. As a result, no edge is added. 

 \begin{equation}
    \label{eq:update_r}
    R(s) \leftarrow \dfrac{R_D(s) + M(s)}{N(s)}
\end{equation}

Second, we update $R(s)$ for visited states---$s_1$ and $s_2$---with Equation \ref{eq:update_r}. We assume the existence of $M(s)$, which is a player model that quantifies player enjoyment of a given segment after gameplay. $N(s)$ tracks the number of times a state was visited---all state counts are initialized to $1$. The value of $N(s)$ is incremented for both $s_1$ and $s_2$, but not $s_3$. By dividing by $N(s)$, we cause reward decay which encourages exploration and disincentivizes repeating states---this may also reduce player fatigue \cite{gravina2019procedural}. 

\begin{equation}
    \label{eq:update_p}
    P(s_a|s,s_a) \leftarrow \dfrac{1+\sum_{x,s_a \in E} \textit{win}(x,s_a)}{1+\sum_{x, s_a \in E}\textit{visits}(x,s_a)}
\end{equation}

Finally, we evaluate the transition probabilities. $E$ is the set of all edges in the graph. We first update the number of times an edge has been used and the number of times the player has beaten the target state. In the example, we update for two edges: $(start, s_1)$ and $(s_1, s_2)$. Visits is incremented for both edges, but only $(start, s_1)$ has the win count incremented. Then we update $P$ with Equation \ref{eq:update_p} for every edge with the target state of $s_1$ or $s_2$. Lastly, we update the probability of entering the death state $P(death|s,s_a) \leftarrow 1-P(s_a|s,s_a)$ for $s_1$ and $s_2$.

After updating the MDP is complete, the system builds a new policy for level assembly.

\subsection{Directors}
\label{ssec:director}

A \textit{director} \cite{yannakakis2012game} adjusts the game to ensure the desired player experience. In this work, we use the term to represent an agent responsible for assembling a level.

\subsubsection{Random}
\label{sssec:random}

\textit{Random} is a baseline director and builds a policy by randomly selecting a neighbor for each state.

\subsubsection{Greedy}

\textit{Greedy} is a baseline director and builds a policy by selecting the neighboring state with the highest reward.

\subsubsection{Policy Iteration (PI)}

The \textit{PI} director uses policy iteration \cite{howard1960dynamic}, a form of ADP, which approximates the utility of states. For every policy update, the utility of all states are set to $0$ and the policy, $\pi_p(s)$, is initialized with random actions. PI then runs \textit{policy evaluation} and \textit{policy improvement} until convergence.\footnote{We found that convergence for an MDP with $9,453$ states takes $<0.5$ seconds running unoptimized Python code.} Policy evaluation updates the utility of each state using equation \ref{eq:policy}, and is run $k$ times---we use $k=20$. $\gamma$ is a discount factor for future rewards. We used $\gamma=0.95$, which we found experimentally by running small trials.

\begin{equation}
    \label{eq:policy}
    U_{i+1}(s) \leftarrow  \sum_{s'} P(s'|s,\pi(s))[R(s') + \gamma U(s')]
\end{equation}

Policy improvement updates the policy to $\pi_{i+1}$ by changing the target state to the state that results in the highest utility according to the updated utility table. If there is a change, the algorithm has not converged and returns to policy improvement. If there is no change, the policy has converged.


\subsubsection{Adaptive Policy Iteration (API)}

We found in case study 2 that PI struggles to adapt when a player persona changes. This motivated the \textit{API} director. The difference between PI and API is that API keeps a running tally of the player's losing streak and removes edges from the start node based on how long the streak is. To show how this tally is used, consider a scenario where the player lost for the first time. API will remove one edge from $start$ to the neighbor with the largest $R_D(s)$. Say that the player loses again. API will remove two edges. This pattern continues till the player wins (in which case the tally is reset) or $start$ only has one edge (in which case the edge is not removed). By doing so, API attempts to avoid nodes that cause continuous player failure.

\subsection{Evaluation}

We have two case studies to evaluate our approach. In the first case study, we model n-gram level generation as an MDP. This is a challenging task because n-grams may not have the full context required to make a level that can be completed by an agent. In the second case study, we use previous work \cite{biemer2021gram,biemer2022linking} that created a graph that connects level segments, and we turn the graph into an MDP. This allows us to always generate completable levels and focus on the player's preference expressed by the player model. Furthermore, the second case study has a much larger MDP in terms of the number of states.

\section{Case Study 1: \textit{Mario} N-Grams}
\XTABLENGramRewards
\XTABLEPlayerPersona
\subsection{Dataset and Graph}

Dahsklog et al. used n-grams to generate full \textit{Mario} levels \cite{dahlskog2014linear} by breaking a \textit{Mario} level into a set of vertical slices (level slices). We build an n-gram ($n=3$) with \textit{Mario} levels from the VGLC \cite{summerville2016vglc}, which are not underground. We use the n-gram to build an MDP where each state represents a prior. There are $513$ states with an average of $1.579$ actions per state and a max of $47$. The designer reward ($R_D(s)$) is $1$ if there is an enemy in the state and $0$ otherwise. 

\subsection{Players}
Assembling levels with an MDP formed from an n-gram does not guarantee a completable level. We use a modified version of the Summerville A* agent \cite{summerville2016vglc} to assess the percent that a level can be completed. As a substitute for a player model ($M(s)$), we define player enjoyment in terms of level slice density (i.e. the number of solid blocks in a column divided by the total number of tiles per column).

\subsection{Evaluation}

We ran each director twenty times on different seeds with fifty levels generated for each run. Fifty levels was chosen because we view it as a small amount in comparison to \textit{Mario}, which has 32 total levels where players will lose multiple times before beating the game. Each level was thirty level slices long. Sample levels can be seen in Appendix B. The results are in Table \ref{XTABLENGramRewards}. In terms of reward, both API and PI are the best performers with an identical average reward and standard deviation. Greedy is the third best performer. Random is the worst. Interestingly, Greedy always produced a completable level. This is the result of getting stuck in a local maxima, which helps explain its low average reward. Next on the list is Random, which performed better in terms of completability than the two ADP methods. API produced levels that were on average more completable than PI. We expected this because API removes potentially problematic states to start from. 

\section{Case Study 2: \textit{Icarus} Segment Generation}
\XTABLEIcarusRewards
\XFIGUREicarusheatmaps
\subsection{Dataset and Graph}

MAP-Elites \cite{mouret2015illuminating} uses a grid formed by tessellating the solution space defined by a set of behavioral characteristics (BC) \cite{smith2010analyzing}. The grid is made up of cells that contain elites, which are the best solutions found that fit the cell's range of BCs. Gram-Elites \cite{biemer2021gram} is an extension of MAP-Elites with a publicly available dataset\footnote{\url{https://github.com/bi3mer/GramElitesData}} of levels for \textit{Kid Icarus}. The BCs used for \textit{Kid Icarus} are density and leniency \cite{smith2010analyzing}. The data set also comes with a graph structure to link level segments \cite{biemer2022linking} that connects neighbors with a directed edge if the concatenated result is beatable and does not contain broken in-game structures. If concatenation fails, a linking algorithm attempts to find a \emph{linker} that satisfies the two requirements. If not possible, there is no edge between the two states. If linking succeeds, a state is created between the two that represents the linker found. There are $9,453$ states and an average of $1.877$ actions per state with a max of $11$. However, there are many linking states and when they are removed we get $1,094$ states with a mean of $8.579$ actions per state and a max of $11$.

Gram-Elites designed the values for each BC to be between zero and one. For states that correspond to a cell, we assign $R_D(s)$ as the mean of the state's BCs. For linking states, we set $R_D(s)$ as the mean of the designer reward for the two linked states---this makes it fairer for the Greedy director. Note that we assume that the larger $R_D$ is for a state, the more difficult the state. The theory is that the more complex the level, the more difficult the level. This may not be true, and we discuss it more in the Conclusion.

There are two final points. First is reward decay. Gram-Elites has multiple elites per cell, which are very similar and may not feel different to the player. We address this by incrementing $N(s)$ for all states in that cell. Second, we gave the example above of generating a level three segments long and counted links as a state. The work on linking level segments guarantees that the resulting level will be completable, but linking states include very few level slices. We address this by not counting them as a full state (e.g. a level supposed to be two states long may have three states where one is a linking state).

\subsection{Players}

We have the guarantee that the level segments combined together result in a level that can be completed by an agent \cite{biemer2021gram}. Instead of evaluating with an agent, we use player proxies \cite{liapis2015procedural, green2018generating}. A player proxy defines the level segments it can beat and how much it enjoyed that level segment. Table \ref{XTABLEPlayerPersona} shows the player proxies we use.

\subsection{Evaluation}

\subsubsection{Segment Generation}

To test segment generation, we ran each director twenty times on different seeds with fifty levels generated for \textit{Icarus} each run. A level is made up of five states; there could be up to four additional linking states. Appendix C shows sample levels. The results are in Table \ref{XTABLEIcarusRewards}. Figure \ref{fig:XFIGUREicarusheatmaps} shows heat maps for three players based on the average number of times a player visited a cell; the highest value of $R_D$ is at the top right of the heat map.

Starting with the average reward, API is the best performer for all player proxies. PI performs second best for all proxies except \GPLEL{}, where it is the worst performer. The reason is shown in the middle plot of Figure \ref{fig:XFIGUREicarusheatmaps}. Both API and PI quickly move to the area with the largest designer reward. Because API reduces failure, it is able to get a larger reward, while PI continually forces the player to try and fail difficult levels. We also see this phenomenon for \BPLHL{}, where PI explores the failure front, whereas API finds the front and moves backward. We see this again for \MPLHD{}.

Moving to percent complete, we can see that the cost for finding levels the player fails is seemingly worse results. However, this is not necessarily problematic. After all, a game where the player never lost would be boring \cite{gonzalez2020finding}. Random is consistently the best for percent complete across all players since it stays near the origin of the graph. We also see one of the failure points for Greedy on \GPLHL{} where it is not able to explore the space enough to find a level that is difficult for the player. API and PI, do not have this problem. 

\subsubsection{Switching Player Proxies}
\XFIGUREswitchingplayerplots

A potential pitfall with systems that adapt to a player is the problem of switching players. A system that fits to a player may only work for that player. When a player switch occurs, we want the director to quickly adapt to the new player. To test this, we ran \GPLHL{} for 35 levels and then switched to \BPLEL{} for 15 levels. These two particular proxies were chosen because the former is an ideal player proxy given the initial reward function, and the latter is not. When the switch occurs, the MDP will favor harder levels that \BPLEL{} cannot beat and does not enjoy. We ran each director twenty times. The directors did not receive notice of the player proxy change. The results are in Figure \ref{fig:XFIGUREswitchingplayerplots}. Initially, API and PI are identical, but there is a slight change when \GPLHL{} starts to fail. When the player proxy switches, we can see that API, Greedy, and PI get a reward of zero. Random never adapted to the first player and its reward goes up since it happened to stay near the easy levels that the new player liked. From level 35 to 50, we can see that Greedy and PI show no signs of adapting. API adapts after about 7 levels and produces playable levels for the new player proxy.

\section{Conclusion}

N-gram generation was difficult for every director because using n-grams to form an MDP is an unideal use case. The first problem is that assembled levels are not guaranteed to be completable. Even once a completable level is found, reward decay encourages the director to find a new level, which results in the user experiencing more levels that are not completable. The second problem is that n-gram generation requires that the policy select many level slices to create a single level, which can result in loops of level slices, as seen in Appendix B for API and PI. This could be addressed by updating $N(s)$ and $R(s)$, building a new policy, and then using that policy to select the next level slice during level assembly, but running such a process over and over again would likely be too slow to be worthwhile. Using level segments helps address the problem of looping because the policy is used much less; looping only becomes a problem if the player makes it to the area with the largest reward, in which case the player has ostensibly beat the game.

Our approach is at its best when assembled levels are guaranteed to be completable. API quickly adapts to the player, but could better explore the solution space. A GLIE scheme \cite{russell_artificial_2009} could address this. 

The assumption that the behavioral characteristics we use are a measure of difficulty is unverified. Work has been done that focuses on measuring difficulty \cite{aponte2011measuring, liapis2013towards, fraser2014methodological}, and is a promising area area for future work.

It may appear that the reliance on linking for levels formed by level segments is a weakness, but this is not the case. Links create the possibility for more potential levels to be made, but are not required. Further, we can look at \textit{Spelunky}, which has a level generator that mixes handmade segments with procedural design \cite{yu2016spelunky}. Instead of a pseudorandom selection of handmade segments, an MDP could be built on top to select segments.

We are interested in two areas for future work. First, we intend to test whether this method is effective as a dynamic difficulty adjustment system with human participants. Second, we want to change our system to use a multi-objective MDP \cite{etessami2007multi, delgrange2020simple} with a finite horizon \cite{sutton2018reinforcement}. The finite horizon will lead to a more accurate utility calculation for the domain of level assembly. Multi-objective optimization reduces the complexity of reward balancing and lets the designer focus more on their game.



\bibliography{refs-manual}

\begin{thebibliography}{31}
\expandafter\ifx\csname natexlab\endcsname\relax\def\natexlab#1{#1}\fi
\providecommand{\url}[1]{\texttt{#1}}
\providecommand{\href}[2]{#2}
\providecommand{\path}[1]{#1}
\providecommand{\DOIprefix}{doi:}
\providecommand{\ArXivprefix}{arXiv:}
\providecommand{\URLprefix}{URL: }
\providecommand{\Pubmedprefix}{pmid:}
\providecommand{\doi}[1]{\href{http://dx.doi.org/#1}{\path{#1}}}
\providecommand{\Pubmed}[1]{\href{pmid:#1}{\path{#1}}}
\providecommand{\bibinfo}[2]{#2}
\ifx\xfnm\relax \def\xfnm[#1]{\unskip,\space#1}\fi
\bibitem[{D'Anastasio(2021)}]{danastasio_2021}
\bibinfo{author}{C.~D'Anastasio}, \bibinfo{title}{Why 2021 was the biggest year
  for the labor movement in games}, \bibinfo{year}{2021}. \URLprefix
  \url{https://www.wired.com/story/2021-biggest-year-labor-movement-video-games/}.
\bibitem[{Shaker et~al.(2016)Shaker, Togelius, and
  Nelson}]{shaker2016procedural}
\bibinfo{author}{N.~Shaker}, \bibinfo{author}{J.~Togelius},
  \bibinfo{author}{M.~J. Nelson}, \bibinfo{title}{Procedural content generation
  in games}, \bibinfo{publisher}{Springer}, \bibinfo{year}{2016}.
\bibitem[{Karth and Smith(2019)}]{karth2019addressing}
\bibinfo{author}{I.~Karth}, \bibinfo{author}{A.~M. Smith},
\newblock \bibinfo{title}{Addressing the fundamental tension of pcgml with
  discriminative learning},
\newblock in: \bibinfo{booktitle}{Proceedings of the 14th International
  Conference on the Foundations of Digital Games}, \bibinfo{year}{2019}, pp.
  \bibinfo{pages}{1--9}.
\bibitem[{Yannakakis et~al.(2013)Yannakakis, Spronck, Loiacono, and
  Andr{\'e}}]{yannakakis2013player}
\bibinfo{author}{G.~N. Yannakakis}, \bibinfo{author}{P.~Spronck},
  \bibinfo{author}{D.~Loiacono}, \bibinfo{author}{E.~Andr{\'e}},
\newblock \bibinfo{title}{Player modeling},
\newblock in: \bibinfo{booktitle}{Artificial and Computational Intelligence in
  Games}, \bibinfo{year}{2013}.
\bibitem[{Hooshyar et~al.(2018)Hooshyar, Yousefi, and Lim}]{hooshyar2018data}
\bibinfo{author}{D.~Hooshyar}, \bibinfo{author}{M.~Yousefi},
  \bibinfo{author}{H.~Lim},
\newblock \bibinfo{title}{Data-driven approaches to game player modeling: a
  systematic literature review},
\newblock \bibinfo{journal}{ACM Computing Surveys (CSUR)} \bibinfo{volume}{50}
  (\bibinfo{year}{2018}) \bibinfo{pages}{1--19}.
\bibitem[{Hunicke(2005)}]{hunicke2005case}
\bibinfo{author}{R.~Hunicke},
\newblock \bibinfo{title}{The case for dynamic difficulty adjustment in games},
\newblock in: \bibinfo{booktitle}{Proceedings of the 2005 ACM SIGCHI
  International Conference on Advances in computer entertainment technology},
  \bibinfo{year}{2005}, pp. \bibinfo{pages}{429--433}.
\bibitem[{Jennings-Teats et~al.(2010)Jennings-Teats, Smith, and
  Wardrip-Fruin}]{jennings2010polymorph}
\bibinfo{author}{M.~Jennings-Teats}, \bibinfo{author}{G.~Smith},
  \bibinfo{author}{N.~Wardrip-Fruin},
\newblock \bibinfo{title}{Polymorph: dynamic difficulty adjustment through
  level generation},
\newblock in: \bibinfo{booktitle}{Proceedings of the 2010 Workshop on
  Procedural Content Generation in Games}, \bibinfo{year}{2010}, pp.
  \bibinfo{pages}{1--4}.
\bibitem[{Liapis et~al.(2015)Liapis, Holmg{\aa}rd, Yannakakis, and
  Togelius}]{liapis2015procedural}
\bibinfo{author}{A.~Liapis}, \bibinfo{author}{C.~Holmg{\aa}rd},
  \bibinfo{author}{G.~N. Yannakakis}, \bibinfo{author}{J.~Togelius},
\newblock \bibinfo{title}{Procedural personas as critics for dungeon
  generation},
\newblock in: \bibinfo{booktitle}{European Conference on the Applications of
  Evolutionary Computation}, \bibinfo{organization}{Springer},
  \bibinfo{year}{2015}, pp. \bibinfo{pages}{331--343}.
\bibitem[{Green et~al.(2018)Green, Khalifa, Barros, Nealen, and
  Togelius}]{green2018generating}
\bibinfo{author}{M.~C. Green}, \bibinfo{author}{A.~Khalifa},
  \bibinfo{author}{G.~A. Barros}, \bibinfo{author}{A.~Nealen},
  \bibinfo{author}{J.~Togelius},
\newblock \bibinfo{title}{Generating levels that teach mechanics},
\newblock in: \bibinfo{booktitle}{Proceedings of the 13th International
  Conference on the Foundations of Digital Games}, \bibinfo{year}{2018}, pp.
  \bibinfo{pages}{1--8}.
\bibitem[{Smith et~al.(2009)Smith, Treanor, Whitehead, and
  Mateas}]{smith2009rhythm}
\bibinfo{author}{G.~Smith}, \bibinfo{author}{M.~Treanor},
  \bibinfo{author}{J.~Whitehead}, \bibinfo{author}{M.~Mateas},
\newblock \bibinfo{title}{Rhythm-based level generation for 2d platformers},
\newblock in: \bibinfo{booktitle}{Proceedings of the 4th international
  Conference on Foundations of Digital Games}, \bibinfo{year}{2009}, pp.
  \bibinfo{pages}{175--182}.
\bibitem[{Gonz{\'a}lez-Duque et~al.(2020)Gonz{\'a}lez-Duque, Palm, Ha, and
  Risi}]{gonzalez2020finding}
\bibinfo{author}{M.~Gonz{\'a}lez-Duque}, \bibinfo{author}{R.~B. Palm},
  \bibinfo{author}{D.~Ha}, \bibinfo{author}{S.~Risi},
\newblock \bibinfo{title}{Finding game levels with the right difficulty in a
  few trials through intelligent trial-and-error},
\newblock in: \bibinfo{booktitle}{2020 IEEE Conference on Games (CoG)},
  \bibinfo{organization}{IEEE}, \bibinfo{year}{2020}, pp.
  \bibinfo{pages}{503--510}.
\bibitem[{Mouret and Clune(2015)}]{mouret2015illuminating}
\bibinfo{author}{J.-B. Mouret}, \bibinfo{author}{J.~Clune},
\newblock \bibinfo{title}{Illuminating search spaces by mapping elites},
\newblock \bibinfo{journal}{arXiv preprint arXiv:1504.04909}
  (\bibinfo{year}{2015}).
\bibitem[{Cully et~al.(2015)Cully, Clune, Tarapore, and
  Mouret}]{cully2015robots}
\bibinfo{author}{A.~Cully}, \bibinfo{author}{J.~Clune},
  \bibinfo{author}{D.~Tarapore}, \bibinfo{author}{J.-B. Mouret},
\newblock \bibinfo{title}{Robots that can adapt like animals},
\newblock \bibinfo{journal}{Nature} \bibinfo{volume}{521}
  (\bibinfo{year}{2015}) \bibinfo{pages}{503--507}.
\bibitem[{Thue and Bulitko(2012)}]{thue2012procedural}
\bibinfo{author}{D.~Thue}, \bibinfo{author}{V.~Bulitko},
\newblock \bibinfo{title}{Procedural game adaptation: Framing experience
  management as changing an {MDP}},
\newblock \bibinfo{journal}{Proceedings of the AAAI Conference on Artificial
  Intelligence and Interactive Digital Entertainment} \bibinfo{volume}{8}
  (\bibinfo{year}{2012}) \bibinfo{pages}{44--50}.
\bibitem[{Shu et~al.(2021)Shu, Liu, and Yannakakis}]{shu2021experience}
\bibinfo{author}{T.~Shu}, \bibinfo{author}{J.~Liu}, \bibinfo{author}{G.~N.
  Yannakakis},
\newblock \bibinfo{title}{Experience-driven pcg via reinforcement learning: A
  super mario bros study},
\newblock in: \bibinfo{booktitle}{2021 IEEE Conference on Games (CoG)},
  \bibinfo{organization}{IEEE}, \bibinfo{year}{2021}, pp.
  \bibinfo{pages}{1--9}.
\bibitem[{Gravina et~al.(2019)Gravina, Khalifa, Liapis, Togelius, and
  Yannakakis}]{gravina2019procedural}
\bibinfo{author}{D.~Gravina}, \bibinfo{author}{A.~Khalifa},
  \bibinfo{author}{A.~Liapis}, \bibinfo{author}{J.~Togelius},
  \bibinfo{author}{G.~N. Yannakakis},
\newblock \bibinfo{title}{Procedural content generation through quality
  diversity},
\newblock in: \bibinfo{booktitle}{2019 IEEE Conference on Games (CoG)},
  \bibinfo{organization}{IEEE}, \bibinfo{year}{2019}, pp.
  \bibinfo{pages}{1--8}.
\bibitem[{Yannakakis(2012)}]{yannakakis2012game}
\bibinfo{author}{G.~N. Yannakakis},
\newblock \bibinfo{title}{Game ai revisited},
\newblock in: \bibinfo{booktitle}{Proceedings of the 9th conference on
  Computing Frontiers}, \bibinfo{year}{2012}, pp. \bibinfo{pages}{285--292}.
\bibitem[{Howard(1960)}]{howard1960dynamic}
\bibinfo{author}{R.~A. Howard}, \bibinfo{title}{Dynamic programming and
  {Markov} processes}, \bibinfo{publisher}{MIT Press}, \bibinfo{year}{1960}.
\bibitem[{Biemer et~al.(2021)Biemer, Hervella, and Cooper}]{biemer2021gram}
\bibinfo{author}{C.~Biemer}, \bibinfo{author}{A.~Hervella},
  \bibinfo{author}{S.~Cooper},
\newblock \bibinfo{title}{Gram-elites: N-gram based quality-diversity search},
\newblock in: \bibinfo{booktitle}{Proceedings of the FDG workshop on Procedural
  Content Generation}, \bibinfo{year}{2021}, pp. \bibinfo{pages}{1--6}.
\bibitem[{Biemer and Cooper(2022)}]{biemer2022linking}
\bibinfo{author}{C.~Biemer}, \bibinfo{author}{S.~Cooper},
\newblock \bibinfo{title}{On linking level segments},
\newblock \bibinfo{journal}{arXiv preprint arXiv:2203.05057}
  (\bibinfo{year}{2022}).
\bibitem[{Dahlskog et~al.(2014)Dahlskog, Togelius, and
  Nelson}]{dahlskog2014linear}
\bibinfo{author}{S.~Dahlskog}, \bibinfo{author}{J.~Togelius},
  \bibinfo{author}{M.~J. Nelson},
\newblock \bibinfo{title}{Linear levels through n-grams},
\newblock in: \bibinfo{booktitle}{Proceedings of the 18th International
  Academic MindTrek Conference: Media Business, Management, Content \&amp;
  Services}, \bibinfo{year}{2014}, pp. \bibinfo{pages}{200--206}.
\bibitem[{Summerville et~al.(2016)Summerville, Snodgrass, Mateas, and
  Ontan{\'o}n}]{summerville2016vglc}
\bibinfo{author}{A.~J. Summerville}, \bibinfo{author}{S.~Snodgrass},
  \bibinfo{author}{M.~Mateas}, \bibinfo{author}{S.~Ontan{\'o}n},
\newblock \bibinfo{title}{The vglc: The video game level corpus},
\newblock \bibinfo{journal}{arXiv preprint arXiv:1606.07487}
  (\bibinfo{year}{2016}).
\bibitem[{Smith and Whitehead(2010)}]{smith2010analyzing}
\bibinfo{author}{G.~Smith}, \bibinfo{author}{J.~Whitehead},
\newblock \bibinfo{title}{Analyzing the expressive range of a level generator},
\newblock in: \bibinfo{booktitle}{Proceedings of the 2010 Workshop on
  Procedural Content Generation in Games}, \bibinfo{year}{2010}, pp.
  \bibinfo{pages}{1--7}.
\bibitem[{Russell and Norvig(2009)}]{russell_artificial_2009}
\bibinfo{author}{S.~Russell}, \bibinfo{author}{P.~Norvig},
  \bibinfo{title}{Artificial intelligence: a modern approach},
  \bibinfo{edition}{3rd edition} ed., \bibinfo{publisher}{Pearson},
  \bibinfo{address}{Upper Saddle River}, \bibinfo{year}{2009}.
\bibitem[{Aponte et~al.(2011)Aponte, Levieux, and Natkin}]{aponte2011measuring}
\bibinfo{author}{M.-V. Aponte}, \bibinfo{author}{G.~Levieux},
  \bibinfo{author}{S.~Natkin},
\newblock \bibinfo{title}{Measuring the level of difficulty in single player
  video games},
\newblock \bibinfo{journal}{Entertainment Computing} \bibinfo{volume}{2}
  (\bibinfo{year}{2011}) \bibinfo{pages}{205--213}.
\bibitem[{Liapis et~al.(2013)Liapis, Yannakakis, and
  Togelius}]{liapis2013towards}
\bibinfo{author}{A.~Liapis}, \bibinfo{author}{G.~Yannakakis},
  \bibinfo{author}{J.~Togelius},
\newblock \bibinfo{title}{Towards a generic method of evaluating game levels},
\newblock in: \bibinfo{booktitle}{Proceedings of the AAAI Conference on
  Artificial Intelligence and Interactive Digital Entertainment},
  \bibinfo{year}{2013}, pp. \bibinfo{pages}{30--36}.
\bibitem[{Fraser et~al.(2014)Fraser, Katchabaw, and
  Mercer}]{fraser2014methodological}
\bibinfo{author}{J.~Fraser}, \bibinfo{author}{M.~Katchabaw},
  \bibinfo{author}{R.~E. Mercer},
\newblock \bibinfo{title}{A methodological approach to identifying and
  quantifying video game difficulty factors},
\newblock \bibinfo{journal}{Entertainment Computing} \bibinfo{volume}{5}
  (\bibinfo{year}{2014}) \bibinfo{pages}{441--449}.
\bibitem[{Yu(2016)}]{yu2016spelunky}
\bibinfo{author}{D.~Yu}, \bibinfo{title}{Spelunky: Boss Fight Books\# 11},
  volume~\bibinfo{volume}{11}, \bibinfo{publisher}{Boss Fight Books},
  \bibinfo{year}{2016}.
\bibitem[{Etessami et~al.(2007)Etessami, Kwiatkowska, Vardi, and
  Yannakakis}]{etessami2007multi}
\bibinfo{author}{K.~Etessami}, \bibinfo{author}{M.~Kwiatkowska},
  \bibinfo{author}{M.~Y. Vardi}, \bibinfo{author}{M.~Yannakakis},
\newblock \bibinfo{title}{Multi-objective model checking of markov decision
  processes},
\newblock in: \bibinfo{booktitle}{International Conference on Tools and
  Algorithms for the Construction and Analysis of Systems},
  \bibinfo{organization}{Springer}, \bibinfo{year}{2007}, pp.
  \bibinfo{pages}{50--65}.
\bibitem[{Delgrange et~al.(2020)Delgrange, Katoen, Quatmann, and
  Randour}]{delgrange2020simple}
\bibinfo{author}{F.~Delgrange}, \bibinfo{author}{J.-P. Katoen},
  \bibinfo{author}{T.~Quatmann}, \bibinfo{author}{M.~Randour},
\newblock \bibinfo{title}{Simple strategies in multi-objective mdps},
\newblock in: \bibinfo{booktitle}{International Conference on Tools and
  Algorithms for the Construction and Analysis of Systems},
  \bibinfo{organization}{Springer}, \bibinfo{year}{2020}, pp.
  \bibinfo{pages}{346--364}.
\bibitem[{Sutton and Barto(2018)}]{sutton2018reinforcement}
\bibinfo{author}{R.~S. Sutton}, \bibinfo{author}{A.~G. Barto},
  \bibinfo{title}{Reinforcement learning: An introduction},
  \bibinfo{publisher}{MIT press}, \bibinfo{year}{2018}.

\end{thebibliography}

\clearpage
\appendix
\begingroup
\setlength{\tabcolsep}{2pt}
\section{Appendix A: Diagram of Level Assembly~System}
\begin{center}
\includegraphics[height=1.5in]{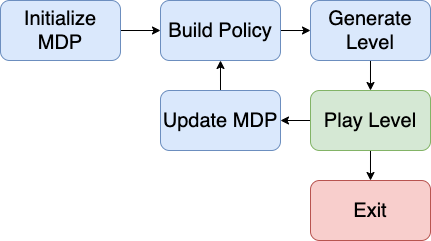}
\end{center}

\section{Appendix B: \textit{Mario} Level Images}
    The first and last completable level assembled by each director for \textit{Mario} n-grams.\\
\begin{center}
    \begin{tabular}{ccc}
        \rotatebox{90}{\strut~~~~~~API} & 
        \includegraphics[height=0.6in]{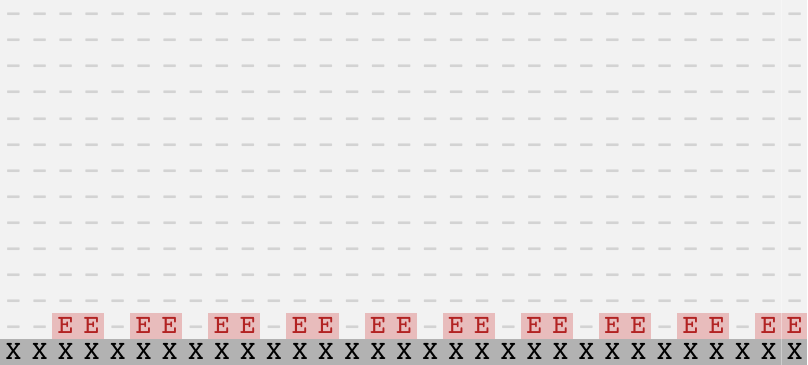} &
        \includegraphics[height=0.6in]{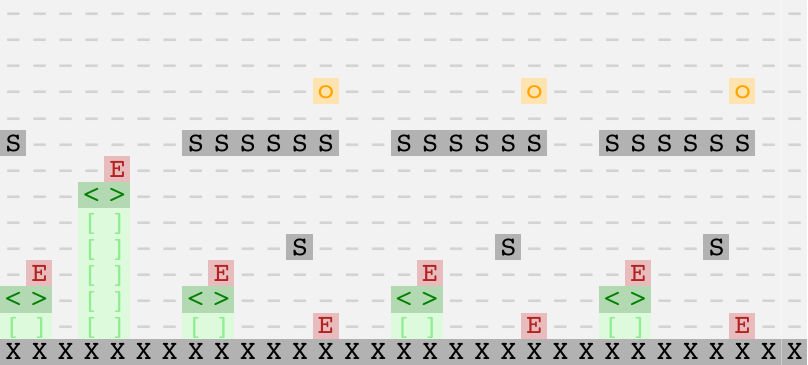}
        \\
        \rotatebox{90}{\strut~~~~~~~PI} &
        \includegraphics[height=0.6in]{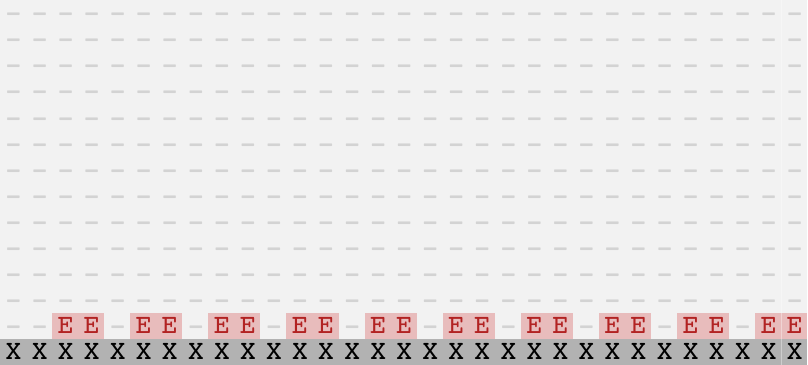}  &
         \includegraphics[height=0.6in]{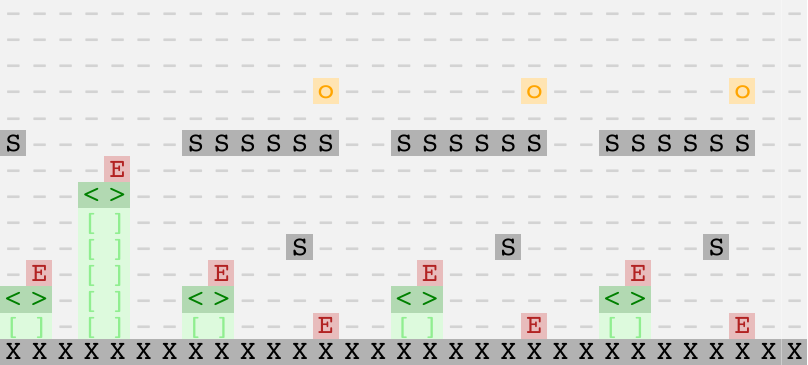}
        \\
        \rotatebox{90}{\strut~~Greedy} &
        \includegraphics[height=0.6in]{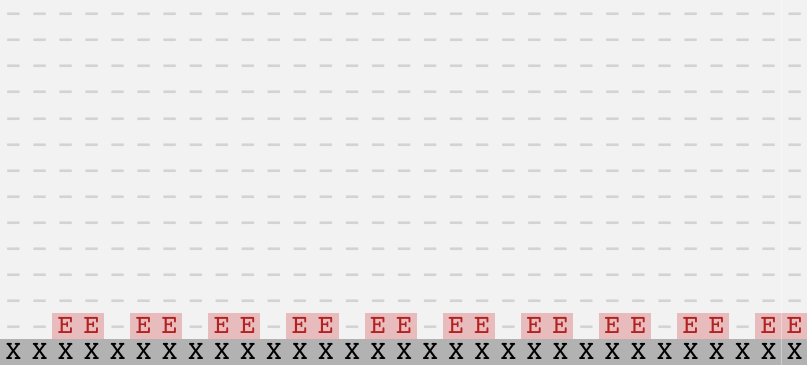} &
        \includegraphics[height=0.6in]{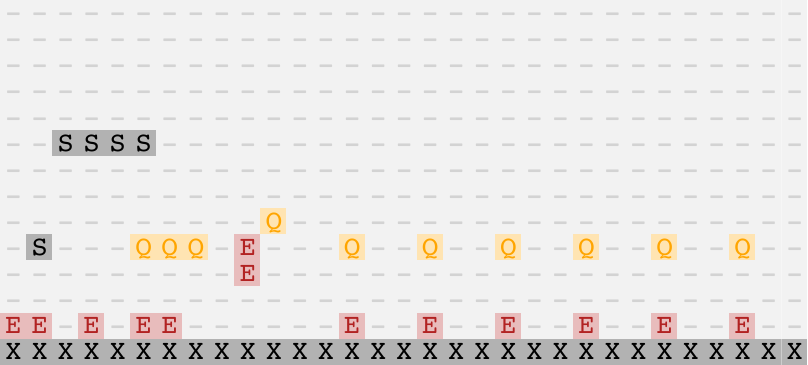}

        \\
        \rotatebox{90}{\strut~~~~Random} &
        \includegraphics[height=0.6in]{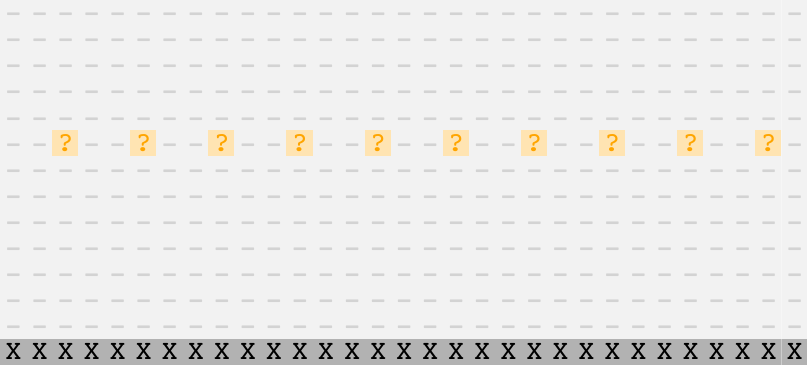} &
        \includegraphics[height=0.6in]{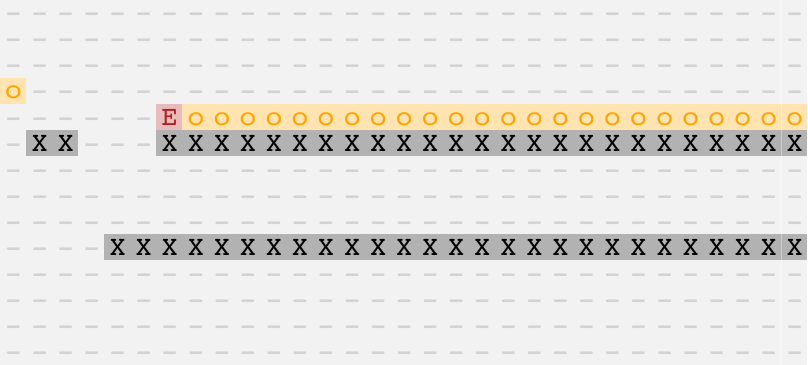}
        \\
        & First Level & Last Level \\
    \end{tabular}
\end{center}
\vfill\eject

\section{Appendix C: \textit{Icarus} Level Images}
    Last completable level assembled for \textit{Icarus} on the last seed for \GPLHL.\\

\begin{center}
    \begin{tabular}{|c|c|c|c|}
    \hline
    \includegraphics[width=0.6in]{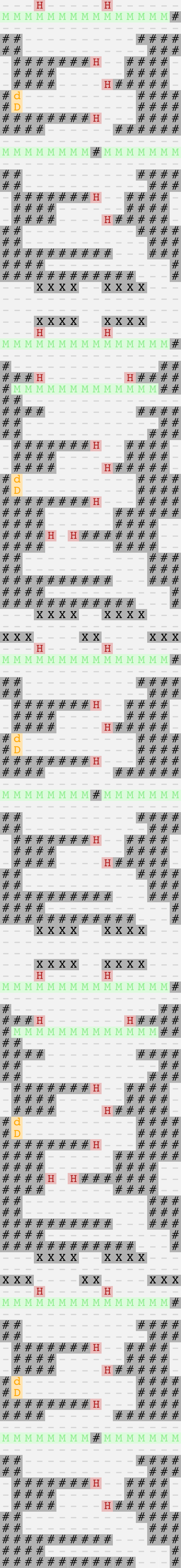} &
    \includegraphics[width=0.6in]{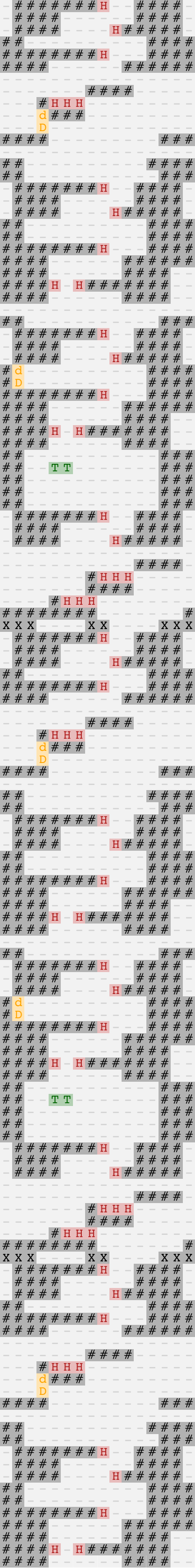} &
    \includegraphics[width=0.6in]{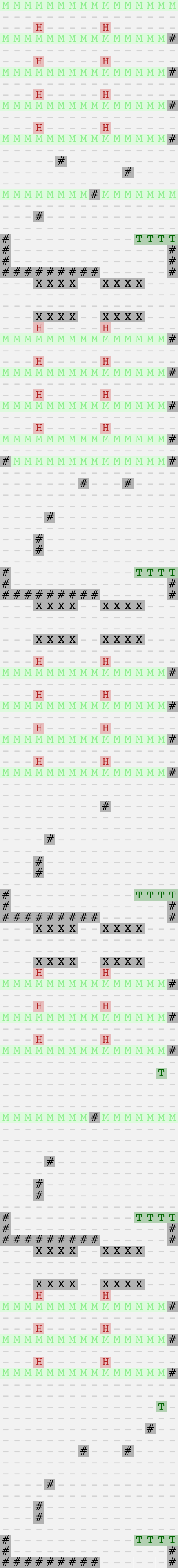} &
    \includegraphics[width=0.6in]{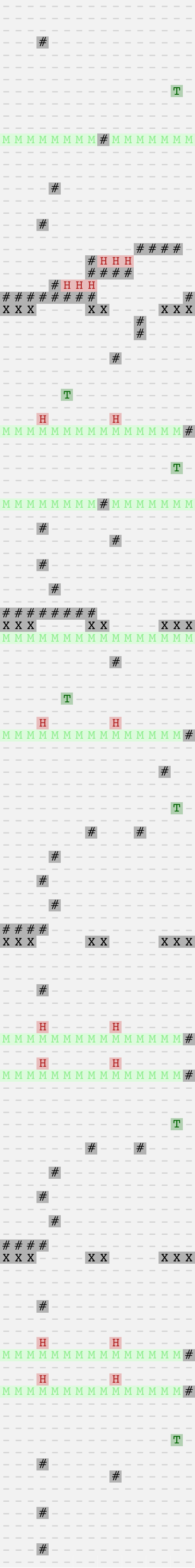} \\
    API & Policy & Greedy & Random\\
    \hline
    \end{tabular}
\end{center}
\vfill

\endgroup

\end{document}